\documentclass{article}

\usepackage[preprint, nonatbib]{neurips_2020}

\usepackage[utf8]{inputenc} 
\usepackage[T1]{fontenc}    
\usepackage{hyperref}       
\usepackage{url}            
\usepackage{booktabs}       
\usepackage{amsfonts}       
\usepackage{nicefrac}       
\usepackage{microtype}      

\usepackage{graphicx}
\usepackage{amsmath}
\usepackage{lipsum}
\usepackage{tikz}
\usepackage[russian,english]{babel}
\usepackage[export]{adjustbox}
\usepackage{float}
\usepackage{makecell}

\title{Interferobot: aligning an optical interferometer \\by a reinforcement learning agent}

\author{
  Dmitry Sorokin$^{1, 4}$ \\
  \texttt{dmitrii.sorokin@phystech.edu} \\
  \And
  Alexander Ulanov$^1$ \\
  \texttt{a.e.ulanov@gmail.com} \\
  \And
  Ekaterina Sazhina$^{1, 4}$ \\
  \texttt{sazhinaekaterina@gmail.com} \\
  \And
  Alexander Lvovsky$^{1,2,3}$ \\
  \texttt{alex.lvovsky@physics.ox.ac.uk} \\ \\
  $^1$\emph{Russian Quantum Center, Moscow, Russia} 
  $^2$\emph{University of Oxford, United Kingdom} \\
  $^3$\emph{P. N. Lebedev Physics Institute, Moscow, Russia} 
  $^4$\emph{Moscow Institute of Physics and Technology}
}

\begin{document}
\maketitle
\begin{abstract}
Limitations in acquiring training data restrict potential applications of deep reinforcement learning (RL) methods to the training of real-world robots. Here we train an RL agent to align a Mach-Zehnder interferometer, which is an essential part of many optical experiments, based on images of interference fringes acquired by a monocular camera. The agent is trained in a simulated environment, without any hand-coded features or a priori information about the physics, and subsequently transferred to a physical interferometer. Thanks to a set of domain randomizations simulating uncertainties in physical measurements, the agent successfully aligns this interferometer without any fine tuning, achieving a performance level of a human expert. 
\end{abstract}


\section{Introduction}

Reinforcement learning is developing explosively nowadays, demonstrating breakthrough results both in computed environments and physical robotics. It achieved above-human performance in Atari video games \cite{Mnih2015, agent57}, defeated the world champion in the game of Go \cite{Silver2017}, and beat the best computer programs in chess and shogi \cite{chess_shogi}. In robotics, RL has been successful in  various manipulation and locomotion tasks 
\cite{dexterous, dexterous_2} such as 
solving Rubik's cube \cite{rubik} and throwing various objects to the goal \cite{tossingbot}. Most of the works considering applications of RL to robotics deal with commercial robotic hands \cite{grasp2vec, replab} or legged robots \cite{minitaur, robel} in well-defined (sometimes changing \cite{qtopt}) environments. Many of these applications have been demonstrated in ``toy'' environments of little practical value. In this paper, we train an RL agent, which we dub \emph{Interferobot}, to solve a commonly occurring practical task in experimental optics: alignment of a Mach-Zehnder interferometer (MZI).

The main bottleneck in the application of RL methods to robotic training is data acquisition. Since RL is heavily based on trial and error, it requires millions of agent-environment interactions to achieve a stable policy. As a result, the training becomes data inefficient and sometimes even dangerous (for example, for autopilots). One way to overcome this problem is to (pre-)train an agent policy in a simulated environment \cite{hierarchical_sim2real}. Training in simulation lets an agent interact with the environment much faster but introduces a domain gap: bias between simulated and real environment. Domain randomization can be used to overcome the domain gap and achieve a stable generic policy \cite{domain_randomization, robust_visual_domain_randomization,cad2rl, dynamics_randomization}, sometimes with additional fine-tuning in a real environment \cite{sim2sim}. However, generalizing this approach to all possible robotic tasks is an open problem.

In present work, we first create a simulator of an MZI and train an agent's policy in this simulated environment based on the observation of images from a monocular camera. The training is implemented without any domain-specific knowledge about the system and hand-coded features. Subsequently, we transfer the trained policy to a physical setup. We add several types of domain randomization in the training procedure in the simulated environment, which allows the transfer to succeed without fine-tuning (in a ``zero-shot'' manner).

Robotic tuning of optical circuits and elements is a dream of many experimental physicists. Tabletop optical experiments require the alignment of tens or even hundreds of different optical elements with micrometric precision. For some experiments, this procedure takes a team of skilled specialists many hours and has to be repeated daily. Fortunately, this alignment is a well-defined task with specified quality metrics and is hence amenable to robotics. However, to our knowledge, control of real-world optical setups by RL agents was not developed before.

\section {Related work}
For relatively simple tasks, it is possible to train a robot's policy directly in the experiment without resorting to a simulated environment \cite{minitaur}, \cite{robel}, \cite{qtopt}. This direct training can be helped with a priori knowledge of the physics and dynamics of the environment. For example in \cite{tossingbot}, the authors trained an agent to throw arbitrary objects into a target box. To achieve that, they embedded a simple ballistic equation into an agent as a starting point and trained it to learn a correction factor. But most often, direct training is not possible due to the environment's complexity. This is especially the case when the robot uses visual information in its policy. In such cases, the pre-training of a policy in a simulated environment is used, followed by transfer and further fine-tuning in the experiment. This method is extensively studied nowadays and different approaches are used. We will name a few. In \cite{sim2sim}, the authors trained a robotic hand to grasp various objects.  They used a generative adversarial network to transform real-world images to look like images from a simulated environment. An alternative approach has been chosen to learn a dexterous in-hand manipulation \cite{cubeopenai} and solve Rubik's cube with a robot hand \cite{rubik}. In these works, a separate neural net was trained in a supervised fashion to extract features from randomized simulated images, which were used as a part of agent's observations. In \cite{cad2rl}, the authors performed a real-world flight of a quadrotor after training only on simulated images. To achieve that, they trained an agent to predict a probability of collisions using heuristics based on free space detection for initialization. In contrast to these references, we trained the agent directly using images, without a priori knowledge about the environment, nor any additional neural nets for image transformation.

Attempts at automated alignment of optical systems included a 1993 study by NASA \cite{NASA}, in which this task was treated as a supervised (rather than reinforcement) learning problem solvable with a multilayer perceptron. The input data consisted of handcrafted features manually extracted from images and were labeled with the actions by a human expert. The overall performance was below that of a human.  Another approach to automated alignment  of a Michelson interferometer with Fabry-Perot cavities was pursued by the VIRGO collaboration in 1997 \cite{VIRGO_ALIGNMENT}. The actions were purely hand-coded involved no neural networks. The method proved to be easy to implement and reliable, but not easily generalizable to interferometers of other types or even with different parameters.  In contrast, we present a general approach that can be applied to a wide range of setups and does not rely on handcrafted features or manually collected datasets. 

\section{Mach-Zehnder interferometer and its simulator}

Interference is a physical phenomenon in which two coherent  waves superimpose to produce oscillations of different amplitudes at different points in space. Dependent on whether the two waves oscillate in or out of phase with respect to each other, the interference can be constructive or destructive. 
In this paper, we consider optical interference of two optical waves --- light beams --- produced by the same laser source. In this case, the interference manifests itself as a pattern of bright and dark fringes visible by eye.

An interferometer is a physical device whose operating principle is based on the interference effect and is used to precisely measure the relative phase difference between two coherent laser beams. It is a primary tool in both academic and applied experimental optics. For example, the Fabry-P\'erot \cite{fabry-perot1899} interferometer is used as a high-resolution optical spectrometer, the Michelson interferometer is the main tool of modern gravitational-wave detectors such as LIGO or VIRGO \cite{LIGO, VIRGO} and also used to precisely measure surface roughness \cite{Kandpal2000}, Sagnac interferometers are used in modern navigation systems, and the  Mach-Zehnder interferometer is a primary tool of modern quantum optical experiments \cite{Ourjoumtsev2006, Sychev2017}.


\begin{figure}
\centering
\includegraphics[width=1.0\linewidth]{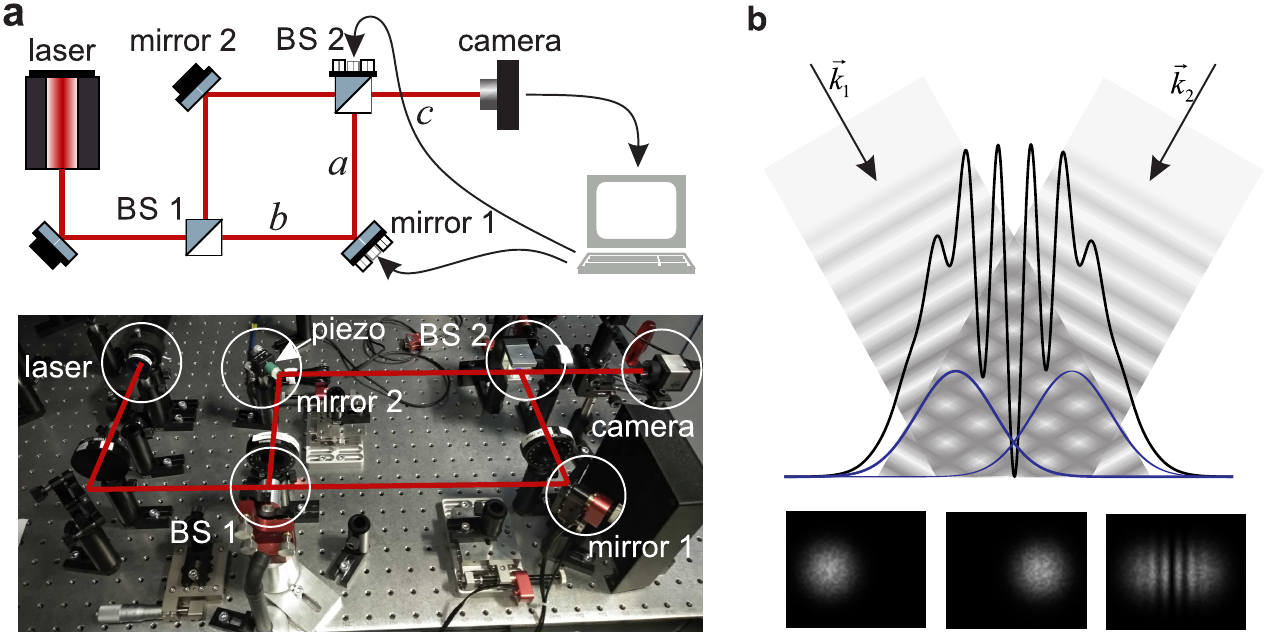}
\caption{a) Conceptual scheme and experimental setup of the Mach-Zehnder interferometer. The interferometer dimensions are listed in Table 1. b) One-dimensional slice of the interferometric pattern produced by two imperfectly overlapping coherent laser beams with wavevectors $\vec k_1$ and $\vec k_2$. Wavefronts are shown via shading; blue lines display the intensities of individual beams (not to scale); black line corresponds to the intensity profile of the resulting picture. Corresponding two-dimensional pictures of individual beams and their interference are displayed at the bottom.}
\label{fig:interf_scheme_and_setup}
\end{figure}

In this paper, we consider a configuration of the MZI shown in Fig. \ref{fig:interf_scheme_and_setup}(a). The laser beam is divided into two at a beam splitter (BS 1).  The two beams go through different interferometer arms and are subsequently recombined at another beam splitter (BS 2). The superimposed beams are detected at a digital camera. Mirror 1 and BS2 are set up in special optomechanical mounts (Newport \cite{newport_mirror}), which can be tilted in two planes either manually using knobs or via an electronic actuator. A camera is used to analyze the resulting interference pattern.

The goal of the alignment is to precisely overlap  the two beams that propagated through the two interferometer arms, to make both the beam centers and wave vectors coincide. The observations that guide the alignment are the images acquired by the camera. A typical interference picture for two misaligned beams is shown in Fig.\ref{fig:interf_scheme_and_setup}(b). In the area where the beams overlap, spatially-dependent fringes appear due to the difference in the beam directions.

\begin{figure}
  \centering
  \includegraphics[width=0.8\linewidth]{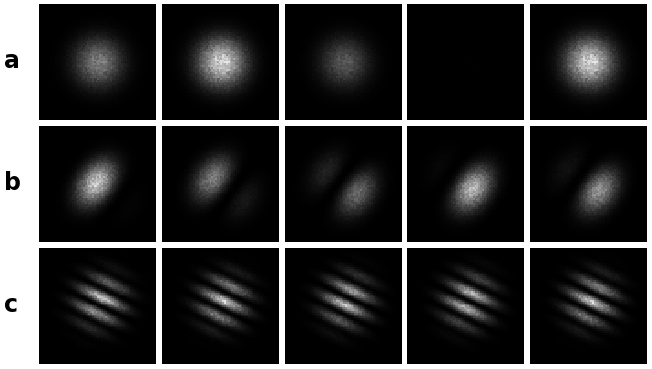}
\caption{Simulated examples of interferometric patterns. (a) Perfectly aligned beams, visibility = 1; (b) Slightly misaligned beams, visibility = 0.3; (c) Strongly misaligned beams, visibility = 0.0026. Pictures from left to right correspond to different relative phases [$\phi_{\mathrm{piezo}}(t)$] between the two interferometer arms induced by the movement of mirror 2 .}
\label{fig:visib_expl}
\end{figure}

An important part of the observation is the temporal dynamics of the  fringes exhibited when  the relative phase between the two beams varied. To this end, mirror 2 is mounted on a piezoelectric transducer which can change the optical path length in the upper arm. Voltage is applied to the piezo in a periodic manner; its peak-to peak amplitude  corresponds to the phase shift of  approximately $2\pi$. For misaligned beams, this motion results in transverse movement of the fringes as seen in Fig.~\ref{fig:visib_expl}(b,c). While the spatial frequency of the fringes is proportional to the absolute  difference between $\vec k_1$ and $\vec k_2$, the direction of their movement provides information about the sign of that difference. For an aligned interferometer, the phase difference of the beams is spatially constant over the camera's sensitive area, so the interference manifests itself as the ``blinking'' of the bright spot [Fig.~\ref{fig:visib_expl}(a)].

To align the interferometer, one needs to interpret the interference patterns and overlap the two beams accurately. It is achieved by tweaking mirror 1 and BS 2 in two orthogonal planes, either manually or via a computer-controlled interface. Despite its apparent simplicity, this alignment is a tedious and time-consuming task. The reason for this is threefold. First, each movement of either mirror 1 or BS 2 lead to simultaneous change both in the position of the lower beam at the camera and its corresponding wave-vector. Hence by trying to improve the spatial overlap between the beams one can lose their parallelism and vice versa. Second, the pattern  measured by the camera is compromised by noise, aberrations and distortions such as dust [Fig.~\ref{fig:rad_fit}(a)]. Third, even advanced optomechanical mounts have limited precision and non-zero hysteresis. 

To train Interferobot, we developed a fast and precise simulator of the MZI. In our model, each laser beam is described by a Gaussian transverse profile with a plane wavefront:
\begin{equation}\label{eq:Exyzt}
E(x,y,z,t) ={\rm Re}\exp\left[\frac{(x-x_0(z))^2+(y-y_0(z))^2}{r^2}\right] \exp(i k_xx+ik_yy+ik_zz+\phi_{\mathrm{piezo}}(t)),    
\end{equation}
where $(x,y,z)$ is the coordinate vector, $(x_0(z),y_0(z))$ the position of the  center of the beam, $\vec{k}$ the wave vector, $r$ the beam radius and $\phi_{\mathrm{piezo}}(t)$ is the phase shift induced by the piezo mirror movement. We neglect the beam divergence due to diffraction and assume paraxial approximation, i.e. $k_z\gg k_x,k_y$, where $z$ is the direction of propagation.   
The intensity of the interferometric picture produced by the two superimposed beams on the camera is then $I(x,y,t)=|E_1(x,y,z,t)+E_2(x,y,z,t)|^2$.

The traditional alignment quality metric is the visibility defined as 
\begin{equation}
    V = \frac{            \max_{t}(I_{\rm tot}) - \min_t(I_{\rm tot})
            } {
                \max_{t}(I_{\rm tot}) + \min_t(I_{\rm tot})
            },
    \label{eq:visib}
\end{equation}
where $I_{\rm tot}(t) = \iint_{-\infty}^{+\infty} I(x, y, t) {\rm d}x{\rm d}y$ is a total optical power incident on the camera; the maximum and minimum are taken over the piezo movement period. The visibility is a real number between 0 and 1. For a fully aligned interferometer [Fig.~\ref{fig:visib_expl}(a)], $\min_t(I_{\rm tot})=0$, so $V=1$, whereas when the interferometer is misaligned [Fig.~\ref{fig:visib_expl}(c)], $\min_t(I_{\rm tot})\approx\max_t(I_{\rm tot})$, so $V\approx 0$.

To simplify the simulation, we assume that  the upper beam propagates exactly along $z$, so $x_0\equiv y_0\equiv k_x\equiv k_y\equiv 0$ for that beam. From Eqs.~\eqref{eq:Exyzt} and \eqref{eq:visib}, we find the visibility to equal 
\begin{equation}
    V = \exp\left(- \frac{x_0^2 + y_0^2}{2 r^2}\right)  \exp\left[- \frac{(k_x^2 + k_y^2) r^2}{8}\right],
    \label{eq:visib_rot}
\end{equation}
where $x_0$ and $y_0$ are the coordinates of the center of the lower beam at the camera and $k_x$ and $k_y$ the components of its wave-vector. 
Assuming that the beam arrives at mirror 1 parallel to the $z$ axis and at the origin of the $(x,y)$ plane, we have 
\begin{equation}
k_{x,y}/k= \alpha_{1x,y} +  \alpha_{2x,y} \textrm{ and }x,y= \alpha_{2x,y} c +  \alpha_{1x,y} (a + c)\label{eq:xykxky}
\end{equation}
where $\alpha_{1x,y}\ll1$ and  $\alpha_{2x,y}\ll1$ are the horizontal and vertical angles by which mirror 1 and BS2, respectively, deflect the beam (with respect to the default $90^\circ$ reflection) while $a$ and $c$ are the interferometer dimensions as shown in Fig.~\ref{fig:interf_scheme_and_setup}(a) and listed in Table 1. We used the fact that, in the paraxial approximation, the horizontal and vertical angles of the beam with respect to the $z$ axis are given by $k_x/k$ and $k_y/k$, respectively.

In the simulation, we represent the camera matrix as a rectangular equidistant grid of $64\times64$ pixels and calculate the  intensity for each pixel. The camera is centered at $(x,y)=(0,0)$, so the beam that has propagated through the upper arm hits the camera in its center. We simulate 16 interferometric patterns distributed in time over a full period of piezo movement corresponding to different $\phi_{\mathrm{piezo}}(t)$. We then calculate the visibility according to Eq.~\eqref{eq:visib}. The fast performance of the simulator is crucial for the training of Interferobot. By using a parallel C++ code, we produce 200 16-frame sets per second. For the training of agents, the simulator has been wrapped into a gym-like interface \cite{brockman2016openai}. 

\begin{table}[ht]
    \caption{Setup parameters. The  interferometer dimensions are given in millimeters, angles in radians.}
    \centering
    \begin{tabular}{|c|c|c|c|c|c|c|c|c|}
    \hline
        parameter& $a$ & $b$ & $c$ & $r$ & $\alpha_{{\rm max}(x,1)}$& $\alpha_{{\rm max}(y,1)}$& $\alpha_{{\rm max}(x,2)}$& $\alpha_{{\rm max}(y,2)}$\\
        \hline
        value & 200 & 300 & 100 & 0.95  & $5.2\cdot10^{-3}$& $3.7\cdot10^{-3}$& $2.6\cdot10^{-3}$& $1.8\cdot10^{-3}$\\
        \hline
    \end{tabular}
    \label{tab:my_label}
\end{table}

\section{Our method}\label{Sec:Method}
\paragraph{States and observations.}
The MZI alignment procedure is treated as a partially observable Markov decision process. The hidden state is the set ($x_0,y_0,k_x,k_y)$, which is not known to Interferobot,
while the observation is the set of 16 consecutive $64\times64$ images taken in one back and forth run of the piezo. Each observation is represented by a (16, 64, 64) tensor.


\paragraph{Actions.}
Interferobot has a multi-discrete action space. Each action is a change in one of the beam deflections $\alpha_{(x,y),(1,2)}$ or $'\mathrm{do\ nothing}'$. We allow three different magnitudes for these changes: $\left[0.01, 0.05, 0.1\right] \times \alpha_{\max}$, where the  maximum angle values $\alpha_{\max}$ are listed in Table 1. It lets the agent choose an appropriate magnitude of action dependent on the degree of misalignment. In total, this results in an action space with the dimension $2\,\mathrm{mounts} \times 2\,\mathrm{dimensions} \times 2\,\mathrm{directions} \times 3\,\mathrm{magnitudes} + 1\,'\mathrm{do\, nothing}' = 25$. 
 
\paragraph{Rewards.}
Visibility itself is not optimal as the reward. First, it is a rather sparse signal, as it exponentially decays with the beam distance and angle as per Eq.~\eqref{eq:visib_rot}. Second, a small difference in the final alignment quality (e.g.~$V = 0.95$ vs. $V = 0.98$) can be of significance for experimental practice. 
Hence we propose the reward signal $R = V - \log(1-V) - 1$. In this way, Interferobot is punished for each additional step, but significantly rewarded for high alignment quality.

\paragraph{Episode and reset.}
The episode length is set to 100 actions. 
At the beginning of each episode, the beam angles are reset to random values within $\pm\alpha_{\max}$. The values of  $\alpha_{\max}$ for each control are chosen to keep both beams within the frame of the camera and to prevent interference pattern features that are small compared to the size of the pixel. 
Specifically, the beam centers can be away from each other by about 3 radii and the interference pattern can have no more than 15 fringes. 

Examples of interference patterns observed (experimentally) after the reset are shown in the  Appendix (Fig.~\ref{fig:interf_patterns}). 

\paragraph{Randomisation.}
To make the learned policy applicable in a physical environment, we applied several domain randomizations inspired by uncertainties in experimental measurements. 
For each episode, we randomly vary the beam radius by $\pm 20\%$, as this parameter is difficult to precisely measure in the experiment. Varying the radius also helps addressing the deviation of the experimental beam profile from precise Gaussian shape.

\begin{figure}
\centering
  \includegraphics[width=0.5\linewidth]{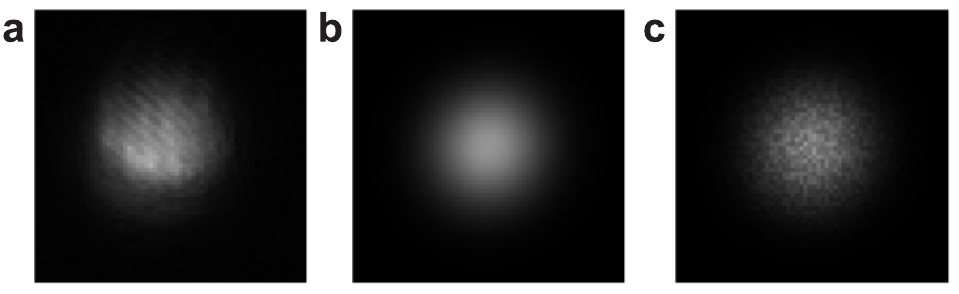}

\caption{Single laser beams: (a) experimental; (b) simulated Gaussian; (c) simulated with noise. In (a), residual fringes are visible due to a resonator effect in the cover glass of the camera.}
\label{fig:rad_fit}
\end{figure}

The following further randomizations are applied once per time step. First, we randomly scaled the brightness by $\pm 30\%$ to model the variable exposure time of the camera. Second, we added $20\%$ of white noise to each pixel to address the camera detector noise as well as the effects of dust and spurious interference  (Fig.~\ref{fig:rad_fit}). The third randomization addresses the challenge of synchronizing the piezo movement with the triggering of the camera. We applied a cyclic shift of simulated interference pattern images within each period of the piezo voltage. In addition, we randomized the ``duty cycle'', i.e.~how the 16 images acquired during the period are distributed between the forward and backward passes of the piezo. The fraction of images acquired during the forward pass was maintained greater than 50\% (at least 9 per period), so Interferobot had enough information to distinguish between the two directions.



\section{Experiments in the simulated environment}
\begin{figure}[ht]
    \centering
    \includegraphics[width=1\linewidth]{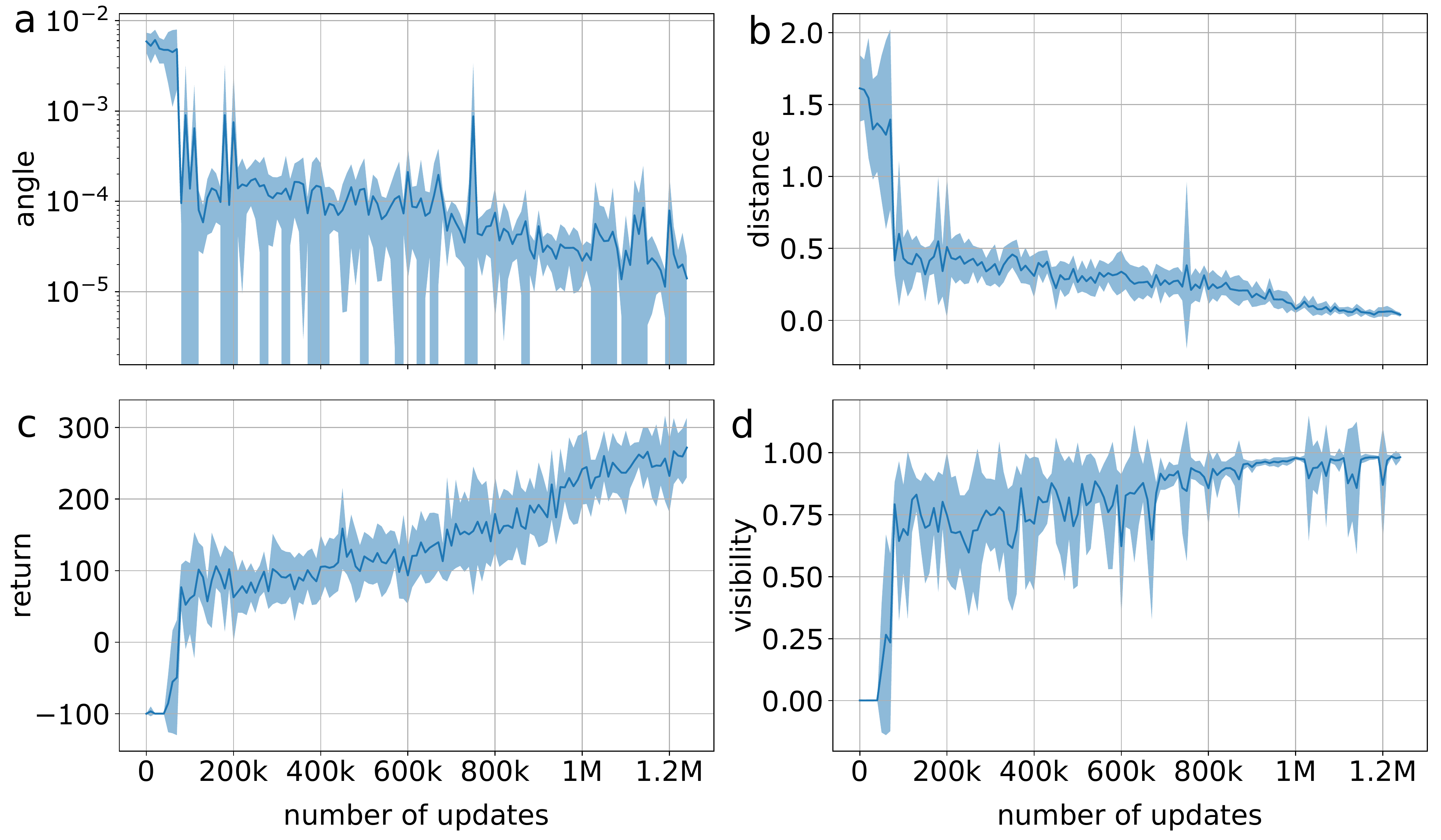}
    \caption{Training in simulation (averaging over 10 training runs). (\textbf{a}) Angle and (\textbf{b}) distance between beams during training, (\textbf{c}) return, (\textbf{d}) visibility. The data in (a), (b) and (d) are for the final steps of each episode. 
    }
    \label{fig:sim_train}
\end{figure}

We train the agent in the simulated environment using the double dueling DQN algorithm \cite{d3qn} with a discount factor $\gamma=0.99$, total number of steps  $5\times10^6$, and replay buffer size $3\times 10^4$. Updates were performed every four steps. The whole training on a NVidia GTX 2060 GPU took about 10 hours. 
The training curves are shown in Fig.~\ref{fig:sim_train}. After the training, both the angle and distance between the beams observed at the and of the episode are below the experimental uncertainty (we recall that the beam radius $r = 0.95$ mm and the angle corresponding to a single fringe is 0.6 mrad). One can also see the benefit of the chosen reward function: while the visibility curve plateaus at the end of the training, the return continues to increase making minute policy improvements manifest.  

{}

\section{Experiments on the physical Mach-Zehnder interferometer}

The experimental MZI is shown in Fig.~\ref{fig:interf_scheme_and_setup}(a). We used a continuous wave laser with the wavelength $\lambda =  635$ nm, Newport Picomotor mirror mounts
an IDS UI-2230SE-M-GL camera with resolution $1600\times1200$ pixels, and acquisition rate 18 frames per second. The acquired images were re-scaled to $64\times64$ pixels. A gym-like interface was created to control the experimental setup by the agent trained in the simulated environment. The action of Interferobot consisted in the corresponding command to the mirror controller. 
The experimental setup generated 0.37 observations per second, corresponding to a single period of the piezo movement.

The trained agent was evaluated automatically without human involvement for 100 episodes of 100 steps each. At the beginning of each episode, the interferometer was reset to a random position as described in the previous section. 

\begin{figure}[ht]
  \includegraphics[width=1\textwidth]{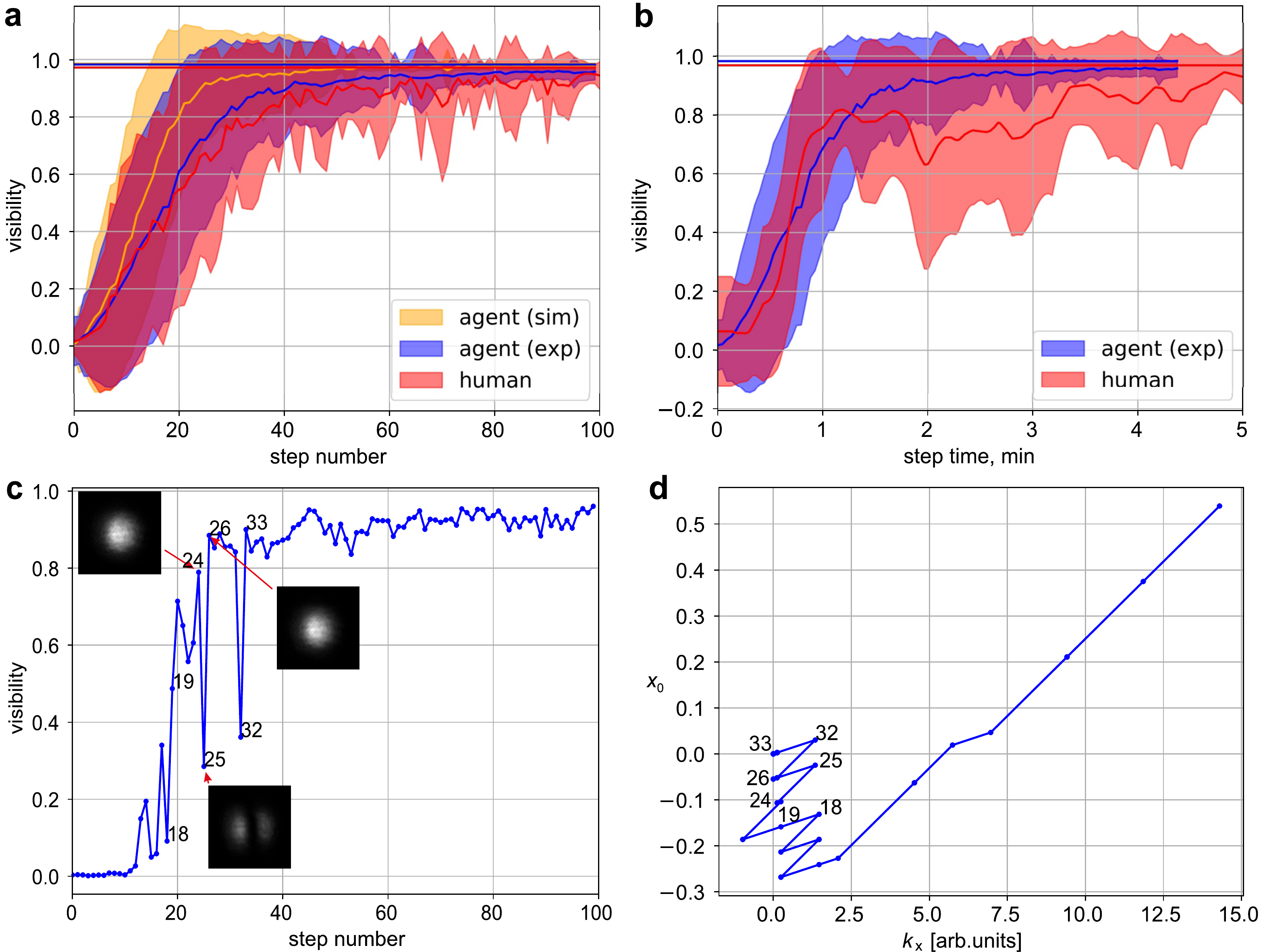}
\caption{Evaluation results.(\textbf{a}) Comparison of Interferobot's performance during an episode in simulation and on the physical setup with a human expert using a keyboard interface. Agent's performance is averaged over 100 evaluations, human's --- over 18. (\textbf{b}) The same but the human manually aligned the MZI by turning knobs. The human's performance is averaged over 16 episodes in this case.  (\textbf{c}) Example visibility evolution in a single experimental episode. The interference pattern photos correspond to steps 24, 25 and 26. (\textbf{d}) Corresponding trajectory in the phase space $(x,k_x)$ shows gradual convergence to the origin.  
}
\label{fig:eval_results}
\end{figure}

In Fig.~\ref{fig:eval_results} we present Interferobot's statistics averaged over all evaluation episodes and compare its performance with that of a skilled human expert. The human expert operated in two settings:  using a keyboard interface with the same set of actions as the agent and  aligning  the interferometer manually using mirror knobs, in line with normal experimental practice. 

Fig.~\ref{fig:eval_results}(a) corresponds to the first setting. In addition to the robotic agent operating in the physical environment, we also show the agent's performance in the simulated environment for reference.  It is seen that the latter surpasses the former due to the domain gap. However, the robotic agent does outperform the human. As an additional comparison metric, we determine the highest visibility obtained in each episode and average it over all episodes. This value  is 0.986 for Interferobot evaluated in the simulation,  0.983 for the  experiment, and 0.972 for the human. Note that, while the performance of the human and robot is comparable when measured in time steps, the human works significantly slower in terms of wall time.


Similar results are obtained in the second setting [Fig.~\ref{fig:eval_results}(b)]. 
Since, in this evaluation, it is difficult to divide the human's procedure into discrete actions, we plot the time, rather than the number of steps, along the horizontal axis. In this setting, the mean highest visibility obtained by the human was 0.968. 
It is seen that the human slightly outperforms the robot in the beginning of each episode, but falls behind when it comes to fine tuning.  
Initially, when the interferometer is significantly misaligned, it is easy for the human to interpret the fringes and obtain the basic alignment in only a few actions. However, when this is achieved,  the human is unable to reliably see a small difference $(x_0,y_0)$ in the beam center positions. To see this difference, the human moves the beams significantly apart, resulting in a temporary visibility drop. Such drop is not present in the first setting [Fig.~\ref{fig:eval_results}(a)] because significant misalignment requires multiple keyboard action, and the human adopts a different fine-tuning strategy: attempting to move the two optical elements by equal angles in opposite directions and observing whether the visibility has improved. 

Interferobot needs to reach, in as few steps as possible, the maximum visibility point, which is located at the origin of the $(x_0,y_0,k_x,k_y)$ space as per Eq.~\eqref{eq:visib_rot}. The agent's actions modify $\alpha_{(1,2),(x,y)}$, and hence, according to Eq.~\eqref{eq:xykxky}, are non-orthogional in that space. This means that the optimal policy will not increase the visibility at every step. Rather, 
it should work in a zig-zag fashion, repeatedly significantly misaligning mirror 1 and then compensating with BS 2 (or vice versa). This behavior, manifested in significant oscillations of the visibility during the alignment, is seen in Fig.~\ref{fig:eval_results}(c,d). Note the difference between these deliberate misalignment actions and those by the human expert mentioned above. While the human misaligns the interferometer in order to better observe the beam positions, Interferobot appears to purposefully make zig-zag steps in order to change $x_0$ or $y_0$ without modifying $k_x$ or $k_y$, respectively.



\section{Ablation study}

\begin{table}[ht]
\caption{Ablation study}
\centering
\begin{tabular}{c|c|c}
\toprule
{} & visibility & return \\
\midrule
All randomizations  & $\textbf{0.96} \pm \textbf{0.02}$ & $\textbf{221} \pm \textbf{54}$ \\
No radius randomization & $0.74 \pm 0.20$ & $85 \pm 69$ \\
No brightness randomization& $0.91 \pm 0.04$ & $178 \pm 39$ \\
No white noise & $0.82 \pm 0.07$ & $129 \pm 43$ \\
No duty cycle and trigger time randomization  &  $0.89 \pm 0.07$ & $200 \pm 42$ \\

\bottomrule
\end{tabular}

\label{tab:abl}
\end{table}






We trained and evaluated four additional agents, in each of which one of the domain randomizations listed in Sec.~\ref{Sec:Method} has been removed. Every agent was evaluated on the physical setup for 100 episodes. The evaluation of each agent took around 6 hours and was fully automated. In table \ref{tab:abl}, we present the mean visibility in the last 20 steps and the mean return, averaged over all episodes. We observe that the most significant benefits are due to the radius randomization and white noise addition.

\section{Summary}

We have shown that an optical interferometer can be successfully aligned by a robotic RL agent
without any prior knowledge of the system's physics. When trained with the randomizations, the agent can be transferred from simulation to an experimental setup without any fine-tuning. The trained agent's performance is comparable with that of a skilled specialist in terms of the achieved visibility, but significantly faster. Our approach can be easily extended to interferometers of various sizes, shapes, and with additional optical elements. In the future, such a method can be applied for fully automated alignment of complex optical setups, which can significantly decrease the amount of routine manual work needed for complex experiments.

\section{Links}
Videos of the interferometer alignment and software are available via link \url{https://github.com/dmitrySorokin/interferobotProject}

\section*{Broader impact}
The direct application of our work is the automation of complex optical experiments. A robotic agent will take over a mundane part  of the experimentalist's work, so scientists can concentrate on generating research ideas and analyzing the scientific results. A potential risk is associated with laser safety, which is a primary concern in optical experiments, as laser beams are able to permanently damage the eyesight. While standard precautions (goggles, barriers, etc.) are normally taken to reduce this risk, additional measures may be required to prevent beams from being misdirected outside the plane of the optical table by the robotic agent. This can be achieved, for example, by limiting the angular range of motorized mirror mounts at the hardware level. 

\begin{ack}

This work was supported by grant UMNIK
\begin{otherlanguage}{russian} 
№ 120ГУЦЭС8-D3/56352
\end{otherlanguage}
from 21.12.2019 and by Russian Science Foundation (19-71-10092). We thank Arsen Kyzhamuratov for help in the human evaluation and Dmitrii Beloborodov for helpful discussions. We also thank Wendelin Boehmer for the feedback on a draft of the manuscript.

\end{ack}

\bibliographystyle{unsrt}


\appendix
\section*{Appendix}

\section{Additional figures}

\begin{figure}[ht]
  \centering
  \includegraphics[width=0.75\textwidth]{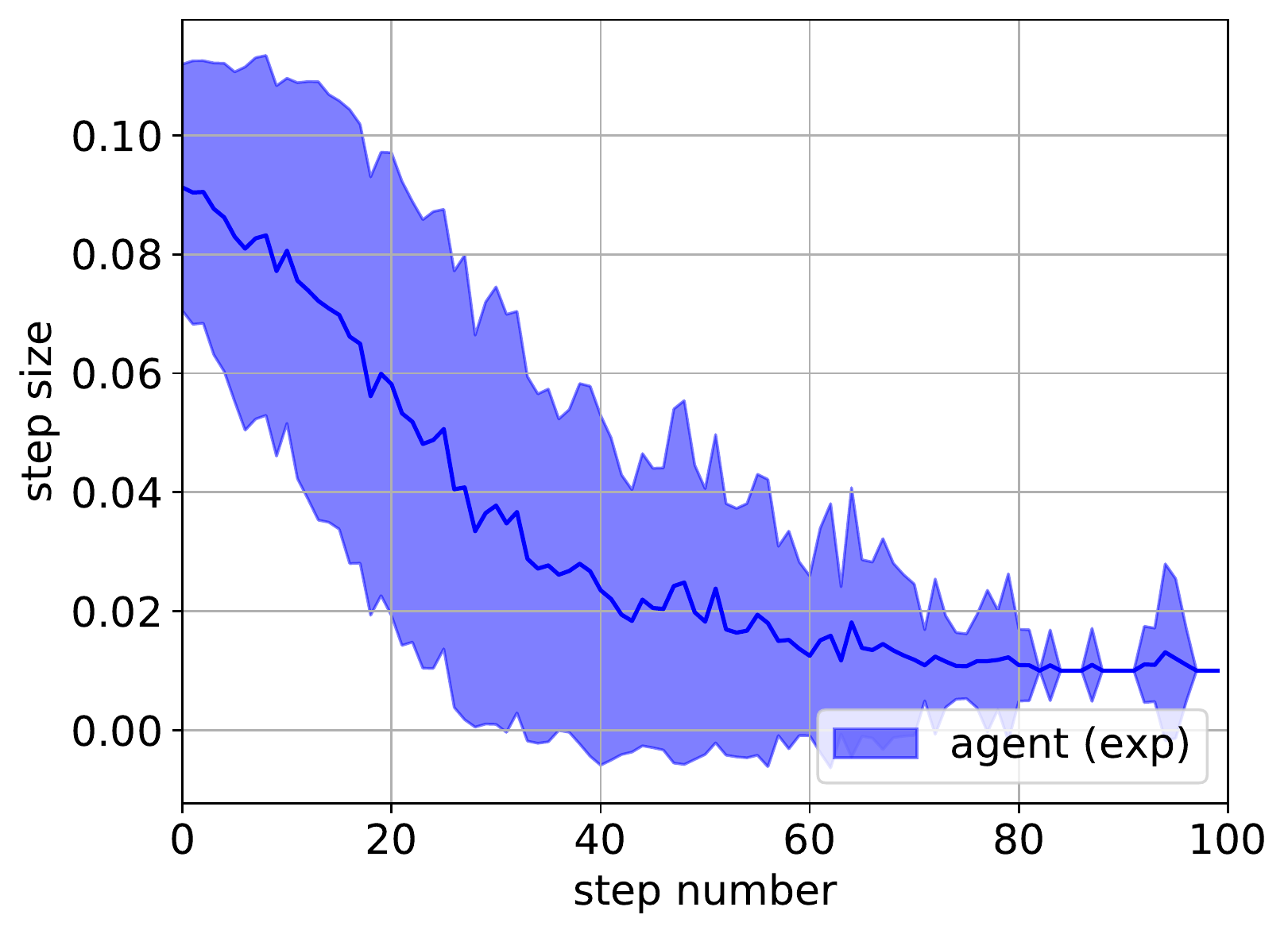}
  \caption{ Agent step length averaged over 100 evaluations.}
  \label{fig:step_size}
\end{figure}

Fig.~\ref{fig:step_size} shows the average action size (in units of $\alpha_{\max}$) during the episode  as a function of step number.  It is seen that in the beginning, when the setup is completely misaligned, Interferobot implements large actions. In the end, the agent performs the smallest actions to fine-tune the interferometer. 

\begin{figure}
\centering

\foreach \x in {
    1, 3, 5, 6, 7, 8, 9, 15, 16, 28, 34, 36, 42, 43, 44, 45, 46, 
    51, 53, 54, 58, 59, 61, 63, 64, 70, 72, 76, 78, 79, 81, 82, 87, 89, 97}
{ 
    \begin{minipage}[b]{0.19\linewidth}
        \centering
        \includegraphics[width=1\linewidth]{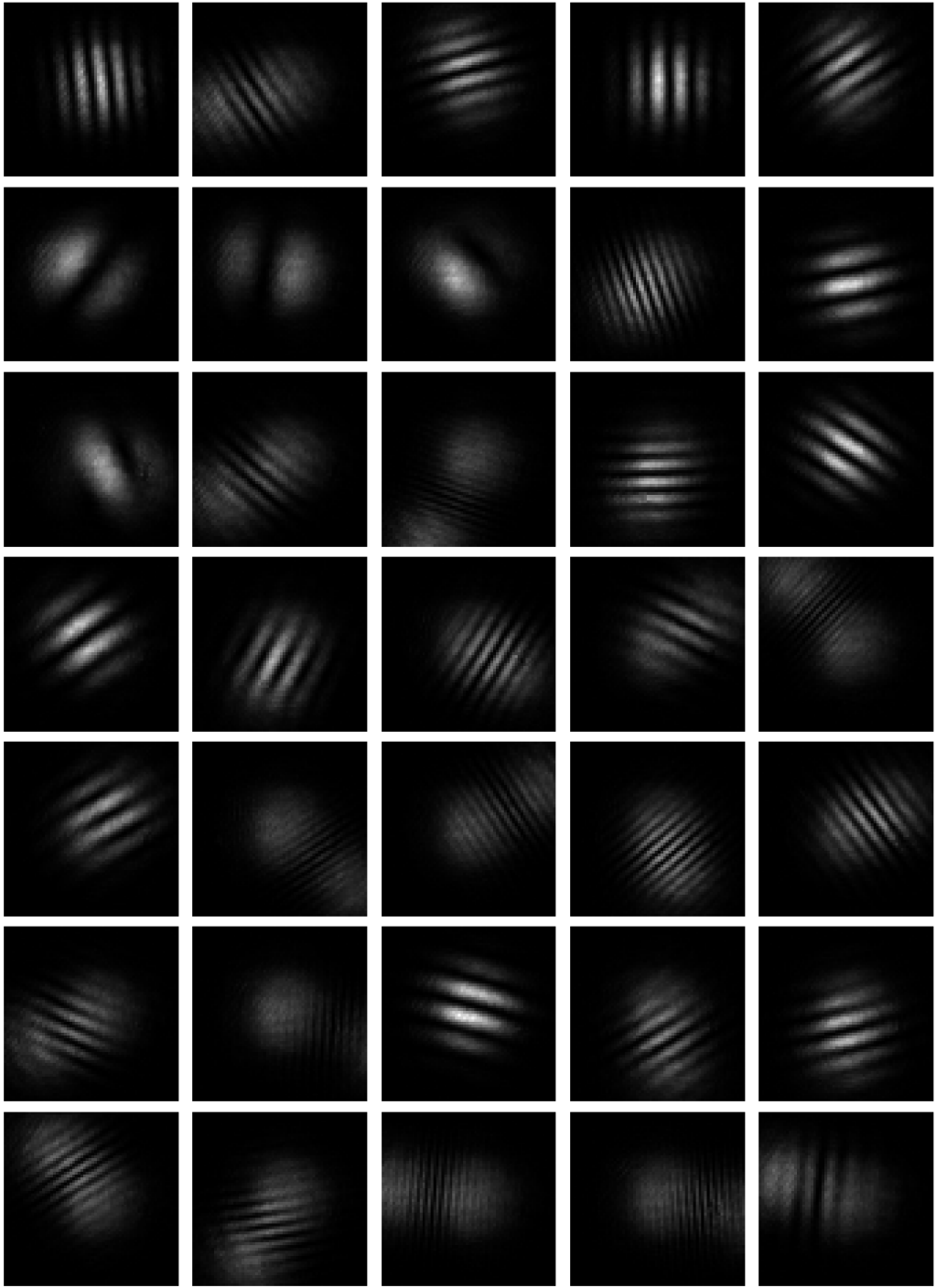}
    \end{minipage}%
}

\caption{Examples of interference patterns experimentally observed after resets.}
\label{fig:interf_patterns}
\end{figure}

Figure \ref{fig:interf_patterns} shows examples of interference patterns observed at the starting point of the episode before the alignment begins.

\end{document}